\documentclass[pmlr]{jmlr} 

\makeatletter
\def\ps@jmlrtps{%
  \let\@mkboth\@gobbletwo
  \def\@oddhead{}%
  \let\@evenhead\@oddhead
  \def\@oddfoot{}%
  \let\@evenfoot\@oddfoot
}
\makeatother

\makeatletter

\ifx\tagdsmode\tagdsproceedings

\else\ifx\tagdsmode\tagdssubmission
  \def\ps@jmlrtps{%
    \let\@mkboth\@gobbletwo
    \def\@oddhead{\scriptsize Under Review at the 2nd Conference on Topology, Algebra, and Geometry in Data Science\hfill}%
    \let\@evenhead\@oddhead
    \def\@oddfoot{}%
    \let\@evenfoot\@oddfoot
  }

\else
  \def\ps@jmlrtps{%
    \let\@mkboth\@gobbletwo
    \def\@oddhead{}%
    \let\@evenhead\@oddhead
    \def\@oddfoot{}%
    \let\@evenfoot\@oddfoot
  }
\fi\fi
\makeatother

\usepackage{longtable}

\usepackage[most]{tcolorbox}

\newtcolorbox{takeawaybox}{
    colback=blue!5,
    colframe=blue!60!black,
    arc=3mm,
    boxrule=0.8pt,
    left=2mm,
    right=2mm,
    top=1mm,
    bottom=1mm,
    fonttitle=\bfseries,
    title=Takeaways
}

\usepackage{booktabs}
\usepackage[load-configurations=version-1]{siunitx} 

\usepackage{enumitem}
\usepackage{wrapfig}

\theorembodyfont{\upshape}
\theoremheaderfont{\scshape}
\theorempostheader{:}
\theoremsep{\newline}

\newcommand{\tagtorch}{\textnormal{\texttt{TAGTorch}}}

\jmlrvolume{334}
\jmlryear{2026}
\jmlrworkshop{Topology, Algebra, and Geometry in Data Science}

\title[TAGTorch: Geometry, Topology, and Symmetry-Aware Machine Learning]{TAGTorch: A PyTorch Library for Geometry, Topology, and Symmetry-Aware Machine Learning}

\author{
\Name{Brendan Kennedy} \Email{}\\
\addr Pacific Northwest National Laboratory
\AND
\Name{Tegan Emerson} \Email{}\\
\addr Pacific Northwest National Laboratory, University of Texas at El Paso
\AND
\Name{Gregory Roek} \Email{}\\
\addr Pacific Northwest National Laboratory
\AND
\Name{Emilie Purvine} \Email{}\\
\addr Pacific Northwest National Laboratory
\AND
\Name{Henry Kvinge} \Email{henry.kvinge@pnnl.gov}\\
\addr University of Washington, Pacific Northwest National Laboratory
}

\begin{document}

\maketitle

\begin{abstract}
Over the last decade, neural networks have been applied to an increasingly diverse range of applications, including data with rich geometric, topological, or symmetry-related structure. As a result, researchers have increasingly drawn inspiration from topology, algebra, and geometry. Despite this rich algorithmic development, the supporting software ecosystem remains fragmented. Many important methods exist only as research prototypes in unmaintained repositories. We address this by introducing \emph{Topology, Algebra, and Geometry Torch (\tagtorch)}, an open-source, PyTorch-based library that unifies tools inspired by topology, algebra, and geometry, including data-preprocessing methods, architectures, training techniques, and model analysis tools. We describe the design philosophy of \tagtorch{} and then discuss its current architecture and capabilities, highlighting areas where it can fill gaps in the current software ecosystem. We conclude with a discussion of our future development priorities for the library.
\end{abstract}

\section{Introduction}
\label{sec:introducton}

\begin{wrapfigure}{r}{0.45\textwidth}
    \centering
    \vspace{-10pt}
    \includegraphics[width=0.43\textwidth]{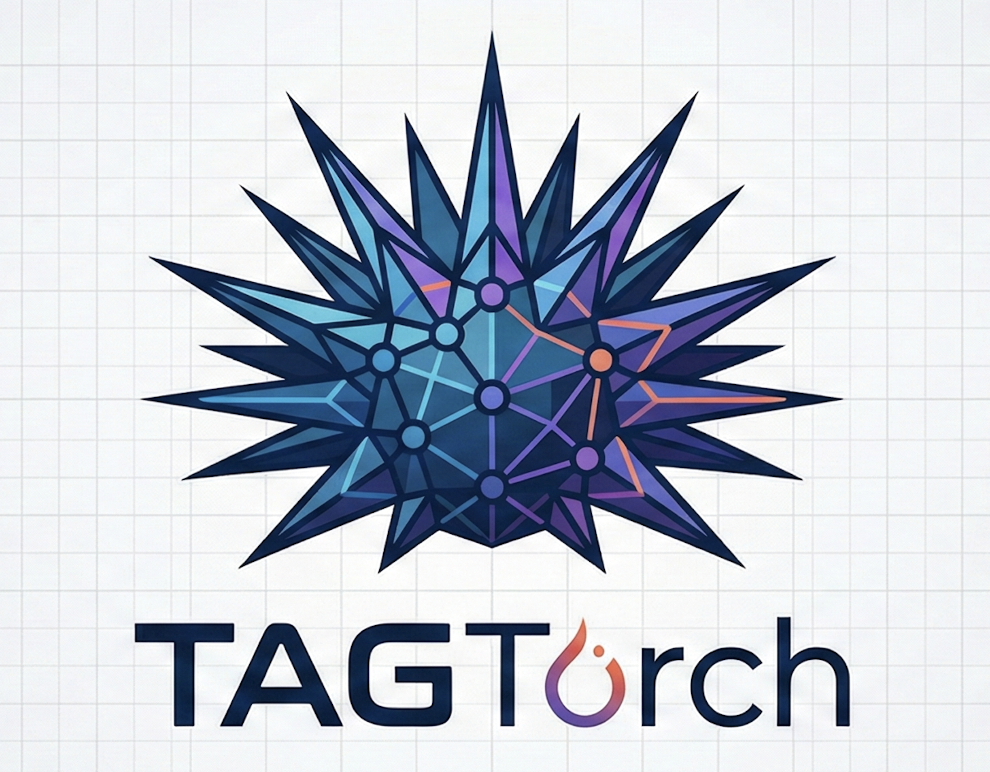}
    \label{fig:tagtorch-logo}
    \vspace{-10pt}
\end{wrapfigure}

The rapid improvement of neural network performance on tasks in natural language and computer vision has triggered an explosion in applications of deep learning to more complex scientific settings. Many of these feature data with complex structure, symmetries, or geometry motivating domain-aware models that benefit from specialized mathematical priors. These can range from group equivariant layer types to input features extracted via topological data analysis. At the same time, the science of deep learning has advanced considerably in the last decade with complex analytical tools now available to understand the behavior and learning dynamics of neural networks. As model internals and the learning process they evolve from are intrinsically mathematical, this field of research has also borrowed heavily from mathematics. Examples include the use of intrinsic dimension statistics to understand the process of learning \citep{ansuini2019intrinsic,joshi2026geometry}, weight symmetries to better understand loss landscape structure \citep{simsek2021geometry,ainsworth2022git,godfrey2022symmetries}, and geometric notions of feature structure such as superposition \citep{elhage2022toy} and the linear representation hypothesis \citep{park2023linear}. 

This transfer of methods and ideas from mathematics to machine learning has been valuable to both communities. Machine learning has received an influx of new ways of looking at models and data and mathematics has been exposed to a host of problems that have the potential to motivate new research directions. But mathematics differs from machine learning in some important ways, one being the role of software. Software remains the bedrock of progress in machine learning. Even work that is primarily theory-based often includes experiments as a way to validate results. Because of this, the success of a machine learning field depends on the software that is available to support it.

What software is available for a researcher interested in using mathematically-inspired tools in AI development? The answer varies considerably based on what area one focuses on. There are some mature, well-maintained libraries with large user-bases. For example, the \texttt{e3nn} library captures a range of methods for building $SO(3)$-equivariant architectures for molecular modeling \citep{geiger2022e3nn}. Another example is \texttt{PyTorch Geometric}, a package aimed at building graph neural networks for learning on structured data \citep{fey2019fast}. While these packages are often effective within the scope that they support they can be hard to work with outside their intended uses. Considerable work might be required to extract a base method (e.g., an equivariance layer construction) from a domain-specific set up (e.g., molecular modeling). This can make it hard to leverage these tools for applications with novel data types or tasks. As another example, chaining together multiple tools in a single workflow (e.g., topological data analysis features with an equivariant architecture) usually requires using several different libraries, leading to potential dependency conflicts. 

To address this, we introduce \emph{Topology, Algebra, and Geometry Torch (\tagtorch)}\footnote{\url{https://github.com/hkvinge/tagtorch}}, a PyTorch-based library that provides implementations of mathematically-inspired methods in deep learning, particularly those based on the fields of topology, algebra, and geometry. The motivating goals of \tagtorch\; are the following. (1) \tagtorch\; is a single entry point to a diverse assortment of TAG-related methods while remaining maximally domain agnostic to accommodate new applications. (2) \tagtorch\; is designed to minimize the barrier to entry for non-mathematicians. (3) \tagtorch\; has a mathematically-informed foundation and structure making it extensible to future developments in the field. (4) Where strong and mature libraries exist, \tagtorch\; uses these rather than replicating capabilities.

In this paper we discuss these and other design goals of \tagtorch, including our strategy of prioritizing one method over another for integration. We also describe our plans to include a strong visualization component to help users without an extensive background build robust intuition for the methods. Finally, we give a brief overview of some of the capabilities that currently exist in \tagtorch. Since \tagtorch\; is still in the early stages of development, we also outline priority features that we are actively planning to integrate in the future.

In summary, our contributions include the following.
\begin{itemize}[itemsep=0pt]
    \item We introduce \tagtorch, a PyTorch-based library that makes mathematical tools for AI development accessible to users beyond experts in machine learning and mathematics.
    \item We describe the design principles that inform \tagtorch\; development.
    \item Finally, we describe the features currently in \tagtorch, as well as features that will be added in the near future.
\end{itemize}

\section{Related work}
\label{sect:related-work}

Ideas from the fields of topology, algebra, and geometry have spurred a range of fruitful research directions in deep learning. Many of these are supported by software libraries that enable efficient experimentation by the community. Examples include \texttt{e3nn} which is specialized to $O(3)$, $SO(3)$, and $E(3)$ symmetries of points in $\mathbb{R}^3$ \citep{geiger2022e3nn}, \texttt{escnn} which captures planar symmetries on grids via $E(n)$-equivariant steerable CNNs \citep{cesa2022a,e2cnn}. Also notable is \texttt{cuEquivariance} which provides low-level tensor operations to support equivariant layers \citep{nvidia-cuequivariance}. Beyond equivariance specifically, \texttt{PyTorch-Geometric (PyG)} is one of the most prominent libraries for geometric deep learning, focusing on learning on graphs \citep{fey2019fast}. Other geometry specific libraries include \texttt{Geomstats} which provides support for several core geometric constructions \citep{JMLR:v21:19-027,miolane2020introduction,guigui2023introduction,le2023parametric,pereira2025learning} and \texttt{Geoopt} which provides manifold optimizers \citep{geoopt2020kochurov}.

The field of topological data analysis (TDA) and topological deep learning has its own ecosystem. Of these \texttt{scikit-tda} is most similar to \tagtorch\; \citep{scikittda2019}. It primarily provides wrappers for other libraries, exposing lots of different packages to users in a convenient format. \texttt{GUDHI} is a well-established, open-source library \citep{maria2014gudhi}. \texttt{Topology Toolkit (TTK)} contains TDA tools for data in dimensions 2 and 3 \citep{tierny2017topology}. Finally, Teaspoon is a TDA package for studying dynamics and time-series data \citep{Khasawneh2025}. 

Lastly, there is an emerging ecosystem of interpretability libraries. Among the most prominent of these is \texttt{TransformerLens}, which contains a range of tools for analysis of transformer-based language models drawn from mechanistic interpretability \citep{nanda2022transformerlens}. Other important libraries include \texttt{NNsight} \citep{fiotto2025nnsight} which aims to enable analysis of model internals at a large scale and \texttt{SAELens} \citep{bloom2024saetrainingcodebase} which is primarily aimed at providing support for sparse autoencoders on language models.

\section{Design Goals}
\label{sec:design-goals}

\tagtorch\; aims to eventually organize and centralize existing tools from the fields of geometric deep learning, topological deep learning and topological data analysis, representation geometry, model symmetries, loss landscape geometry, and mathematical interpretability techniques. We particularly hope to implement missing links and engineer high-level flows to reduce the barrier to entry into the world of TAG and deep learning. Our strategy blends an appreciation for the wide array of exciting research in the space, as well as closing gaps and building bridges between communities and sub-fields. By building this package, we aim to start a community-wide initiative for contributing new methods to a centralized project, which will amplify individual research efforts and ease direct comparisons, evaluations, and applications of methods in the TAG toolbox.

\begin{figure}
\centering
\includegraphics[width=0.99\textwidth]{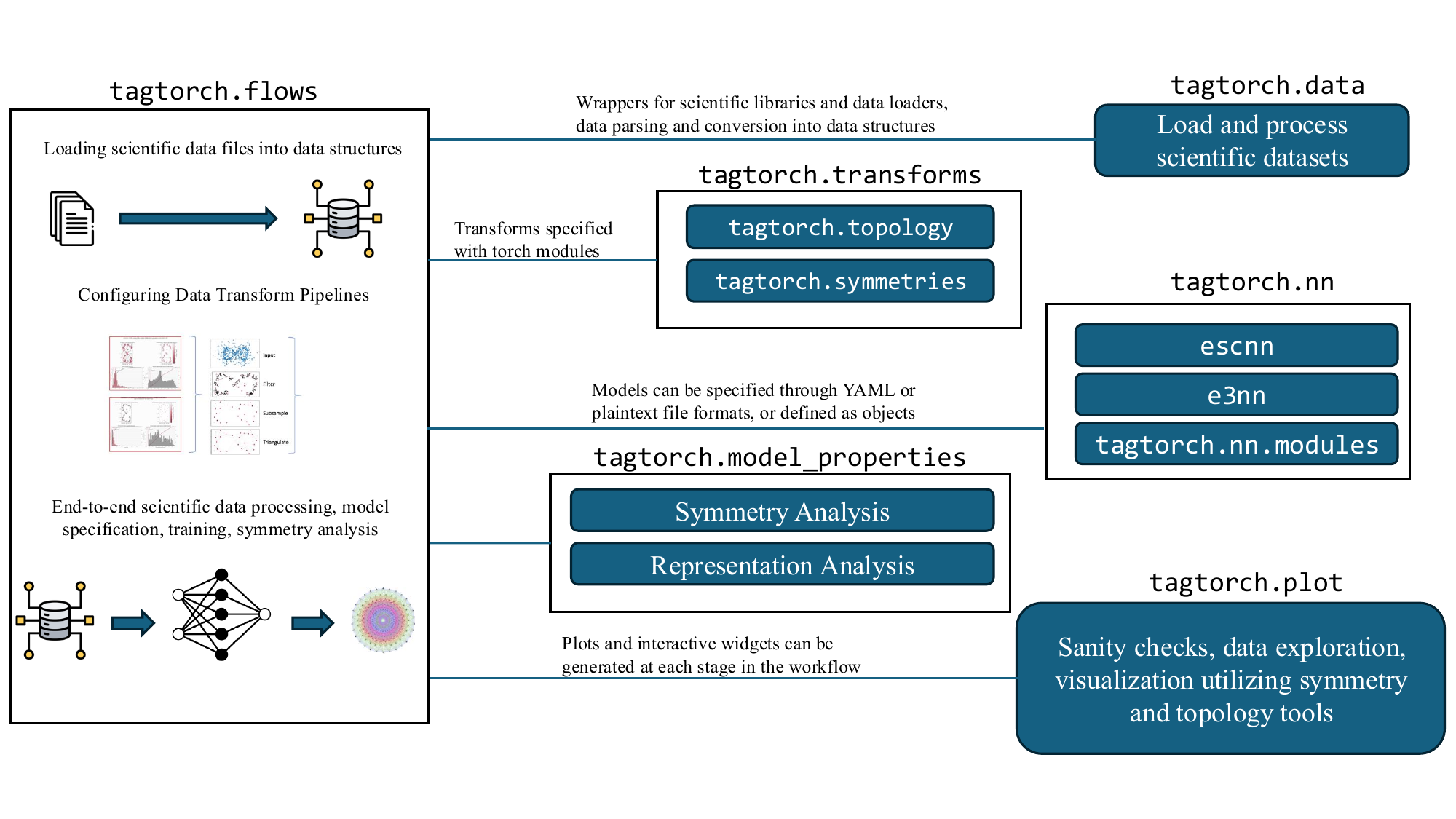}
\caption{A diagram of the current structure of \tagtorch. This will be a starting point upon which we will build out specific capabilities.}
\label{fig:tagtorch-diagram}
\end{figure}

\subsection{Organizing and centralizing existing tools}

As a community, there are many scattered implementations of algorithms and specialized tools that all operate on universal mathematical constructions. However, reconciling these implementations runs into the issue that there is significant heterogeneity in verbiage and formalisms used across software packages. Some of this comes from the fact that the implementations originate from different disciplines (e.g., mathematics vs. physics). Some arises from the applications that the implementations are geared toward. In either case, the consequence is it can take some work to translate between tools, even when they operate within an identical mathematical framework. 

To make this concrete, consider the example of representation theory, which serves as the foundation for equivariant neural networks. Each group has a well-defined set of isomorphism classes of irreducible linear representations \citep{serre1977linear,fulton2013representation}. These serve as universal building blocks for all symmetries of the group. They also provide a unifying grammar for understanding and comparing different approaches to equivariant machine learning that leverage that group. However, it is very hard to see these connections at the level of software.

Beyond the fact that the current ecosystem fails to elevate structure that could help unify different implementations, the field also suffers from maladies common to any growing research area. This includes situations where the only implementations of algorithms are often one-off, unmaintained research repositories. These sometimes allow replication, but they usually lack the engineering necessary to provide a general capability to a larger community.

In this space, \tagtorch\; aims to amplify existing works and contextualize them within a common framework. This will involve:
\begin{itemize}[itemsep=0pt]
    \item Implementing methods within a well-engineered software architecture, which standardizes access points to methods, uses abstractions and object-oriented programming, and adds extensive testing for robustness and improvements to efficiency.
    \item Defining abstractions such that the myriad formalisms used across subfields and software packages can be interacted with in a standardized way (e.g., a universal language for irreducible representations).
    \item Connecting methods to the PyTorch ecosystem. This unifying backend brings methods into contact with sophisticated training capabilities and optimizations, evaluation utilities, and methods from other deep learning disciplines such as computer vision or natural language processing.
\end{itemize}

\subsection{Implementation of New Methods}

In cases where methods lack accepted implementations from the community, our goal will be to add them to \tagtorch. We will aim to do this in a way that aligns them with competing approaches. For example, when a new tool for analyzing representation geometry appears, it will be added alongside existing representation geometry tools aligning available hyperparameters, assumptions on input data, etc. as much as reasonably possible. By aiming for implementation alignment, we hope to make it easy to compare competing methods without the friction of having to integrate a new code base with different assumptions and formalisms.

\subsection{Reducing the Barrier to Entry}

Centralizing and rounding out the TAG toolbox only goes so far in terms of making a tool that can be of use to a wider community. We aim for \tagtorch\; to occupy a shared space between fundamental and applied science, connecting real data and use cases to researchers in deep learning and mathematics. This is a symbiotic relationship: fundamental science often develops in a closed community, leveraging data that is useful for innovation but untethered from the domain sciences which are, in theory, benefiting from this novel research; meanwhile, domain scientists may not be aware of relevant methods coming from the machine learning community. This divergence is compounded by the lack of unifying software.

To reduce the barrier to entry for domain scientists into the world of TAG-flavored AI tools, and in turn drive AI research with real applications, \tagtorch\; provides high-level workflows that help ``connect the dots'' between distinct phases in the research cycle, from data processing and transformation, visualization, building mathematically-aware architectures, and training and evaluating models informed by the geometry of data.

\subsection{Extensibility}

Ultimately the success of \tagtorch\; will largely depend on community engagement, both in terms of feedback on the package, but also direct contributions. We have made two major design choices to try and encourage this. First, we have implemented object-oriented design principles in various modules in the package. With this, contributors can add new functionality to the package with minimal changes required to the codebase,  adhering to the encapsulation principle of software architecture. Designed class hierarchies in \tagtorch\; include: data loaders and processors, which currently are wrappers around other libraries' data loading capabilities; torch data transforms using methods from topological and symmetry-based analysis; equivariant torch architectures; and implementations of groups and group actions for symmetry analysis. Each of these can be integrated into stable, package-level flows that implement multiple steps in analysis, modeling, and evaluation.

\section{Design and Architecture}
\label{sec:architecture}

\subsection{A standardized API for data transforms}

Transforms are central in modern mathematics. As category theory tells us, we should focus on the morphisms, not the objects. Transforms are also central to the world of traditional deep learning and PyTorch, deep learning essentially being the process of composing many simple transformations together. Beyond that, familiar transforms like augmentation, mutation, and sampling are integrated into workflows in popular libraries, including `torchvision` \citep{marcel2010torchvision} and `torch-geometric` \citep{fey2019fast}. PyTorch-compatible transform layers can be directly integrated into data loaders and training and evaluation processes. 

In \tagtorch, we use the notion of a transform to incorporate many central capabilities within a single framework. Currently, the library includes topological transforms and data symmetries, but we will add weight symmetries and representation statistics as well. Workflows in TAG require non-trivial transformations among data types, particularly when working with graphical data. \tagtorch\; combines native implementations of common graph utilities with imports of standard libraries.

\subsection{Visualization and interactive utilities}

It is non-trivial to interpret data and train models without strong intuition. This is particularly true when techniques are built on methods from graduate level mathematics. For instance, equivariant architectures are generally harder to interpret than their non-equivariant counterparts, especially if one does not have some basic intuition about representation theory. 

In some cases, specific types of visualizations can be leveraged to aid in intuition building. In \tagtorch, some methods will be supported by custom visualization procedures. One feature that is currently in development is a visualization pipeline that helps a user understand parameter choice in persistence homology (Figure \ref{SM:fig:hyperparam-tuning-widget} in the Appendix). In this particular instance, the user can generate visualizations that show the impact of different hyperparameter choices (specifically density floor and $\epsilon$-net which filter out points and maximum distance and maximum number of neighbors which filter out edges).

\section{Capabilities}
\label{sec:capabilities}

In this section we outline \tagtorch's current functionality as well as functionality in development. 

\subsection{Symmetry and Equivariance}

We have taken a mathematics-first approach to building symmetry and equivariance capabilities in \tagtorch. This means starting with groups, then defining actions and linear representations of these groups on specific datatypes, and then moving to downstream functionality. By building up from algebraic foundations, we make the package more extensible. We have protocols that define a class interface to build a new action or representation for an existing group. These are all currently in the \texttt{tagtorch.symmetry} module. This framework also helps us to connect different packages that use the same underlying groups and representations but use different syntax to invoke them. For example, \texttt{e3nn} and \texttt{escnn} both utilize the same underlying mathematics, but you would not know this from looking at the way the code is set up. We hope to smooth out these differences. 

Beyond these foundations, we have started by including basic functionality such as incorporating symmetry actions into a dataloader for convenient augmentation. We have also elevated some of the best resources for equivariant architectures, including \texttt{e3nn} \citep{geiger2022e3nn} and \texttt{escnn} \citep{e2cnn,cesa2022a}, making these easily callable from within \tagtorch. Other equivariant architectures that we consider essential are those associated with sets, such as DeepSets \citep{zaheer2017deep} and Set Transformer \citep{lee2019set}. We plan to also include other prominent methods in equivariant learning that are not architectural, such as frame averaging methods \citep{puny2021frame,ma2024canonicalization} and methods which promote a softer inductive bias toward equivariance \citep{finzi2021residual}.

\subsection{Topology}

Our ambition is to include two types of topology-inspired tools in \tagtorch. The first are tools coming from topological data analysis. These involve extraction of topological statistics from data either for input features to a downstream neural network or as a tool for analyzing training or inference data. \tagtorch\; already includes the Euler characteristic transform (ECT) \citep{turner2014persistent} and basic persistent homology capabilities. We aim to eventually include skeletonization via Morse theory approaches \citep{gyulassy2008efficient}, Mapper \citep{singh2007topological}, standard topological vectorizations, zig-zag persistence \citep{carlsson2010zigzag}, and sheaf-theoretic methods \citep{curry2014sheaves}. In the longer term we also plan to include support for topology-aware layers \citep{papamarkou2024position} which is a different flavor of topology-inspired tool.

As noted in Section \ref{sect:related-work}, there are several mature Python-based software packages for TDA. Our goal with \tagtorch{} is to utilize these where possible, including extra functionality where it will make integration into PyTorch-based workflows easier. In certain cases, where no maintained tool is available, we plan to provide our own implementations.

Beyond the algorithms themselves, we also provide resources for hyperparameter optimization. This includes identifying the specific parameters most relevant to a given dataset, maximizing the information extracted under a restricted compute budget, and rapidly visualising simplicial complexes for qualitative analysis/validation  (identifying data points excluded as outliers, included/excluded edges and triangles, etc.). 

\subsection{Model Properties}

One fundamental area where development has been limited is our \tagtorch\texttt{.model\_properties} modules which will include mathematical tools for better understanding neural networks. This will include functions to estimate intrinsic dimension \citep{camastra2016intrinsic}, functions for analyzing representation geometry \citep{lin2024topology}, functions for representation similarity comparison \citep{klabunde2025similarity}, and functions to measure other mathematical properties of a model. At present we have a function for calculating the group empirical equivariance deviation \citep{kvinge2022ways} which measures the extent to which a neural network is equivariant.  

\section{Conclusion}

In addition to providing new, cross-disciplinary capabilities, \tagtorch\; is a vehicle for interdisciplinary cohesion and the sharing of methods, utilities, and common practices. Much as other disciplines in machine learning have benefited from centralized software and model sharing software (e.g., torch hub\footnote{\url{https://pytorch.org/hub/}}, HuggingFace\footnote{\url{https://huggingface.co/}}, or torchvision\footnote{\url{https://docs.pytorch.org/vision/stable/index.html}}), the communities focused on developing and applying TAG methods will benefit from having comprehensive method implementations integrated into the PyTorch ecosystem, continuously updated by the package maintainers and supported by contributions from the community.

\tagtorch\; is in its early stages, and will continue evolving in terms of content and user flows but maintain the initial structures and architecture described here. These structures will create simple, accessible means for community members to contribute to the package, taking advantage of abstractions to add support for new models, data structures and transforms, tools for model property analysis, and visual tools. Completed implementations are highlighted by topological and symmetry-based transforms, a unifying implementation and formalism for implementing symmetries in PyTorch, interactive tools for visualizing data and applying persistent homology, and wrapper functions and classes around existing python libraries for equivariant architectures.

At the same time, \tagtorch\; is actively being fleshed out beyond the initial architecture and structural design. Active areas of development include: (1) implementations of new equivariant architectures beyond the wrapper functionality we provide for \texttt{e3nn} and \texttt{escnn}; (2) interactive Graph User Interfaces (GUI) for managing human-in-the-loop workflows, such as hyperparameter tuning; and (3) specific flows for data loading and processing and model development. By open-sourcing \tagtorch, we invite the community to contribute to these and other areas in the package, adding diversity to the present implementations and providing a platform for others to use and extend their work.


\bibliography{main}

@article{ansuini2019intrinsic,
  title={Intrinsic dimension of data representations in deep neural networks},
  author={Ansuini, Alessio and Laio, Alessandro and Macke, Jakob H and Zoccolan, Davide},
  journal={Advances in Neural Information Processing Systems},
  volume={32},
  year={2019}
}

@article{kvinge2022ways,
  title={In what ways are deep neural networks invariant and how should we measure this?},
  author={Kvinge, Henry and Emerson, Tegan and Jorgenson, Grayson and Vasquez, Scott and Doster, Tim and Lew, Jesse},
  journal={Advances in Neural Information Processing Systems},
  volume={35},
  pages={32816--32829},
  year={2022}
}

@article{joshi2026geometry,
  title={Geometry of decision making in language models},
  author={Joshi, Abhinav and Bhatt, Divyanshu and Modi, Ashutosh},
  journal={Advances in Neural Information Processing Systems},
  volume={38},
  pages={144139--144196},
  year={2026}
}

@inproceedings{simsek2021geometry,
  title={Geometry of the loss landscape in overparameterized neural networks: Symmetries and invariances},
  author={Simsek, Berfin and Ged, Fran{\c{c}}ois and Jacot, Arthur and Spadaro, Francesco and Hongler, Cl{\'e}ment and Gerstner, Wulfram and Brea, Johanni},
  booktitle={International Conference on Machine Learning},
  pages={9722--9732},
  year={2021},
  organization={PMLR}
}

@article{godfrey2022symmetries,
  title={On the symmetries of deep learning models and their internal representations},
  author={Godfrey, Charles and Brown, Davis and Emerson, Tegan and Kvinge, Henry},
  journal={Advances in Neural Information Processing Systems},
  volume={35},
  pages={11893--11905},
  year={2022}
}

@article{ainsworth2022git,
  title={Git re-basin: Merging models modulo permutation symmetries},
  author={Ainsworth, Samuel K and Hayase, Jonathan and Srinivasa, Siddhartha},
  journal={arXiv preprint arXiv:2209.04836},
  year={2022}
}

@article{elhage2022toy,
  title={Toy models of superposition},
  author={Elhage, Nelson and Hume, Tristan and Olsson, Catherine and Schiefer, Nicholas and Henighan, Tom and Kravec, Shauna and Hatfield-Dodds, Zac and Lasenby, Robert and Drain, Dawn and Chen, Carol and others},
  journal={arXiv preprint arXiv:2209.10652},
  year={2022}
}

@article{park2023linear,
  title={The linear representation hypothesis and the geometry of large language models},
  author={Park, Kiho and Choe, Yo Joong and Veitch, Victor},
  journal={arXiv preprint arXiv:2311.03658},
  year={2023}
}

@article{geiger2022e3nn,
  title={e3nn: Euclidean neural networks},
  author={Geiger, Mario and Smidt, Tess},
  journal={arXiv preprint arXiv:2207.09453},
  year={2022}
}

@article{fey2019fast,
  title={Fast graph representation learning with PyTorch Geometric},
  author={Fey, Matthias and Lenssen, Jan Eric},
  journal={arXiv preprint arXiv:1903.02428},
  year={2019}
}

@book{serre1977linear,
  title={Linear representations of finite groups},
  author={Serre, Jean-Pierre and others},
  volume={42},
  year={1977},
  publisher={Springer}
}

@book{fulton2013representation,
  title={Representation theory: a first course},
  author={Fulton, William and Harris, Joe},
  year={2013},
  publisher={Springer Science \& Business Media}
}

@inproceedings{marcel2010torchvision,
  title={Torchvision the machine-vision package of torch},
  author={Marcel, S{\'e}bastien and Rodriguez, Yann},
  booktitle={Proceedings of the 18th ACM international conference on Multimedia},
  pages={1485--1488},
  year={2010}
}

@inproceedings{cesa2022a,
        title={A Program to Build {E(N)}-Equivariant Steerable {CNN}s },
        author={Gabriele Cesa and Leon Lang and Maurice Weiler},
        booktitle={International Conference on Learning Representations},
        year={2022},
        url={https://openreview.net/forum?id=WE4qe9xlnQw}
}

@inproceedings{e2cnn,
       title={{General E(2)-Equivariant Steerable CNNs}},
       author={Weiler, Maurice and Cesa, Gabriele},
       booktitle={Conference on Neural Information Processing Systems (NeurIPS)},
       year={2019},
       url={https://arxiv.org/abs/1911.08251}
}

@software{nvidia-cuequivariance,
  author = {{NVIDIA Corporation}},
  title = {cuEquivariance},
  year = {2025},
  url = {https://github.com/NVIDIA/cuEquivariance}
}

@article{JMLR:v21:19-027,
    author  = {Nina Miolane and Nicolas Guigui and Alice Le Brigant and Johan Mathe and Benjamin Hou and Yann Thanwerdas and Stefan Heyder and Olivier Peltre and Niklas Koep and Hadi Zaatiti and Hatem Hajri and Yann Cabanes and Thomas Gerald and Paul Chauchat and Christian Shewmake and Daniel Brooks and Bernhard Kainz and Claire Donnat and Susan Holmes and Xavier Pennec},
    title   = {Geomstats:  A Python Package for Riemannian Geometry in Machine Learning},
    journal = {Journal of Machine Learning Research},
    year    = {2020},
    volume  = {21},
    number  = {223},
    pages   = {1-9},
    url     = {http://jmlr.org/papers/v21/19-027.html}
}

@article{miolane2020introduction,
    title     = {Introduction to geometric learning in python with geomstats},
    author    = {Miolane, Nina and Guigui, Nicolas and Zaatiti, Hadi and Shewmake, Christian and Hajri, Hatem and Brooks, Daniel and Le Brigant, Alice and Mathe, Johan and Hou, Benjamin and Thanwerdas, Yann and others},
    journal   = {SciPy 2020-19th Python in Science Conference},
    pages     = {48--57},
    year      = {2020},
    url       = {https://proceedings.scipy.org/articles/Majora-342d178e-007.pdf}
}

@article{guigui2023introduction,
    author  = {Guigui, Nicolas and Miolane, Nina and Pennec, Xavier and others},
    title   = {Introduction to riemannian geometry and geometric statistics: from basic theory to implementation with geomstats},
    journal = {Foundations and Trends in Machine Learning},
    volume  = {16},
    number  = {3},
    pages   = {329--493},
    year    = {2023},
    publisher = {Now Publishers, Inc.},
    url     = {https://www.nowpublishers.com/article/Details/MAL-098},
}

@article{le2023parametric,
    author  = {Le Brigant, Alice and Deschamps, Jules and Collas, Antoine and Miolane, Nina},
    title   = {Parametric information geometry with the package Geomstats},
    journal = {ACM Transactions on Mathematical Software},
    volume  = {49},
    number  = {4},
    pages   = {1--26},
    year    = {2023},
    publisher = {ACM New York, NY}
}

@article{pereira2025learning,
    title   = {Learning from landmarks, curves, surfaces, and shapes in Geomstats},
    author  = {Pereira, Lu{\'\i}s F and Brigant, Alice Le and Myers, Adele and Hartman, Emmanuel and Khan, Amil and Tuerkoen, Malik and Dold, Trey and Gu, Mengyang and Su{\'a}rez-Serrato, Pablo and Miolane, Nina},
    journal = {ACM Transactions on Mathematical Software},
    year    = {2025},
    publisher = {ACM New York, NY}
}

@misc{geoopt2020kochurov,
    title={Geoopt: Riemannian Optimization in PyTorch},
    author={Max Kochurov and Rasul Karimov and Serge Kozlukov},
    year={2020},
    eprint={2005.02819},
    archivePrefix={arXiv},
    primaryClass={cs.CG}
}

@misc{scikittda2019,
    author       = {Nathaniel Saul and Chris Tralie},
    title        = {Scikit-TDA: Topological Data Analysis for Python},
    year         = 2019,
    doi          = {10.5281/zenodo.2533369},
    url          = {https://doi.org/10.5281/zenodo.2533369}
}

@inproceedings{maria2014gudhi,
  title={The gudhi library: Simplicial complexes and persistent homology},
  author={Maria, Cl{\'e}ment and Boissonnat, Jean-Daniel and Glisse, Marc and Yvinec, Mariette},
  booktitle={International congress on mathematical software},
  pages={167--174},
  year={2014},
  organization={Springer}
}

@article{tierny2017topology,
  title={The topology toolkit},
  author={Tierny, Julien and Favelier, Guillaume and Levine, Joshua A and Gueunet, Charles and Michaux, Michael},
  journal={IEEE transactions on visualization and computer graphics},
  volume={24},
  number={1},
  pages={832--842},
  year={2017},
  publisher={IEEE}
}

@Article{Khasawneh2025,
    author    = {Khasawneh, Firas A. and Munch, Elizabeth and Barnes, Danielle and Chumley, Max M. and Güzel, İsmail and Myers, Audun D. and Tanweer, Sunia and Tymochko, Sarah and Yesilli, Melih},
    journal   = {Journal of Open Source Software},
    title     = {Teaspoon: A Python Package for Topological Signal Processing},
    year      = {2025},
    issn      = {2475-9066},
    month     = mar,
    number    = {107},
    pages     = {7243},
    volume    = {10},
    doi       = {10.21105/joss.07243},
    publisher = {The Open Journal},
}

@misc{nanda2022transformerlens,
    title = {TransformerLens},
    author = {Neel Nanda and Joseph Bloom},
    year = {2022},
    howpublished = {\url{https://github.com/TransformerLensOrg/TransformerLens}},
}

@inproceedings{fiotto2025nnsight,
  title={NNsight and NDIF: Democratizing access to open-weight foundation model internals},
  author={Fiotto-Kaufman, Jaden and Loftus, Alexander and Todd, Eric and Brinkmann, Jannik and Pal, Koyena and Troitskii, Dmitrii and Ripa, Michael and Belfki, Adam and Rager, Can and Juang, Caden and others},
  booktitle={International Conference on Learning Representations},
  volume={2025},
  pages={92337--92370},
  year={2025}
}

@misc{bloom2024saetrainingcodebase,
   title = {SAELens},
   author = {Bloom, Joseph and Tigges, Curt and Duong, Anthony and Chanin, David},
   year = {2024},
   howpublished = {\url{https://github.com/decoderesearch/SAELens}},
}

@article{zaheer2017deep,
  title={Deep sets},
  author={Zaheer, Manzil and Kottur, Satwik and Ravanbakhsh, Siamak and Poczos, Barnabas and Salakhutdinov, Russ R and Smola, Alexander J},
  journal={Advances in neural information processing systems},
  volume={30},
  year={2017}
}

@inproceedings{lee2019set,
  title={Set transformer: A framework for attention-based permutation-invariant neural networks},
  author={Lee, Juho and Lee, Yoonho and Kim, Jungtaek and Kosiorek, Adam and Choi, Seungjin and Teh, Yee Whye},
  booktitle={International conference on machine learning},
  pages={3744--3753},
  year={2019},
  organization={PMLR}
}

@article{puny2021frame,
  title={Frame averaging for invariant and equivariant network design},
  author={Puny, Omri and Atzmon, Matan and Ben-Hamu, Heli and Misra, Ishan and Grover, Aditya and Smith, Edward J and Lipman, Yaron},
  journal={arXiv preprint arXiv:2110.03336},
  year={2021}
}

@article{ma2024canonicalization,
  title={A canonicalization perspective on invariant and equivariant learning},
  author={Ma, George and Wang, Yifei and Lim, Derek and Jegelka, Stefanie and Wang, Yisen},
  journal={Advances in Neural Information Processing Systems},
  volume={37},
  pages={60936--60979},
  year={2024}
}

@article{finzi2021residual,
  title={Residual pathway priors for soft equivariance constraints},
  author={Finzi, Marc and Benton, Gregory and Wilson, Andrew G},
  journal={Advances in Neural Information Processing Systems},
  volume={34},
  pages={30037--30049},
  year={2021}
}

@article{turner2014persistent,
  title={Persistent homology transform for modeling shapes and surfaces},
  author={Turner, Katharine and Mukherjee, Sayan and Boyer, Doug M},
  journal={Information and Inference: A Journal of the IMA},
  volume={3},
  number={4},
  pages={310--344},
  year={2014},
  publisher={Oxford University Press}
}

@article{gyulassy2008efficient,
  title={Efficient Computation of Morse-Smale Complexes for Three-Dimensional Scalar Functions},
  author={Gyulassy, Attila and Natarajan, Vijay and Pascucci, Valerio and Bremer, Peer-Timo and Hamann, Bernd},
  journal={IEEE Transactions on Visualization and Computer Graphics},
  volume={13},
  number={6},
  pages={1440--1447},
  year={2007}
}

@inproceedings{singh2007topological,
  author = {Singh, Gurjeet and M{\'e}moli, Facundo and Carlsson, Gunnar},
  title = {Topological Methods for the Analysis of High Dimensional Data Sets and 3D Object Recognition},
  booktitle = {Symposium on Point-Based Graphics},
  pages = {91--100},
  year = {2007},
  publisher = {Eurographics Association},
  doi = {10.2312/SPBG/SPBG07/091-100}
}

@article{carlsson2010zigzag,
  author  = {Carlsson, Gunnar and de Silva, Vin},
  title   = {Zigzag Persistence},
  journal = {Foundations of Computational Mathematics},
  volume  = {10},
  number  = {4},
  pages    = {367--405},
  year     = {2010},
  doi      = {10.1007/s10208-010-9066-0}
}

@book{curry2014sheaves,
  title={Sheaves, Cosheaves and Applications},
  author={Curry, Justin},
  year={2014},
  publisher={University of Pennsylvania}
}

@article{papamarkou2024position,
  title={Position paper: Challenges and opportunities in topological deep learning},
  author={Papamarkou, Theodore and Birdal, Tolga and Bronstein, Michael and Carlsson, Gunnar and Curry, Justin and Gao, Yue and Hajij, Mustafa and Kwitt, Roland and Li{\`o}, Pietro and Di Lorenzo, Paolo and others},
  journal={arXiv preprint arXiv:2402.08871},
  volume={2},
  year={2024}
}

@article{camastra2016intrinsic,
  title={Intrinsic dimension estimation: Advances and open problems},
  author={Camastra, Francesco and Staiano, Antonino},
  journal={Information Sciences},
  volume={328},
  pages={26--41},
  year={2016},
  publisher={Elsevier}
}

@article{lin2024topology,
  title={The topology and geometry of neural representations},
  author={Lin, Baihan and Kriegeskorte, Nikolaus},
  journal={Proceedings of the National Academy of Sciences},
  volume={121},
  number={42},
  pages={e2317881121},
  year={2024},
  publisher={National Academy of Sciences}
}

@article{klabunde2025similarity,
  title={Similarity of neural network models: A survey of functional and representational measures},
  author={Klabunde, Max and Schumacher, Tobias and Strohmaier, Markus and Lemmerich, Florian},
  journal={ACM Computing Surveys},
  volume={57},
  number={9},
  pages={1--52},
  year={2025},
  publisher={ACM New York, NY}
}

\appendix

\section{Workflows and Interactive Applications}\label{apd:workflows}

\subsubsection{Interactive widgets for hyperparameter optimization}

In many cases, data visualization provides valuable signal that can help a model developer evaluate goodness of fit, guide the design of data augmentation or transformation, and model design. In \tagtorch, we are developing a suite of end-to-end methods for tying data visualization to model development. In particular, the initial release of \tagtorch\; features an interactive Graphical User Interface for selecting hyperparameters when applying persistent homology.

\begin{figure}
    \centering
    \includegraphics[scale=0.5]{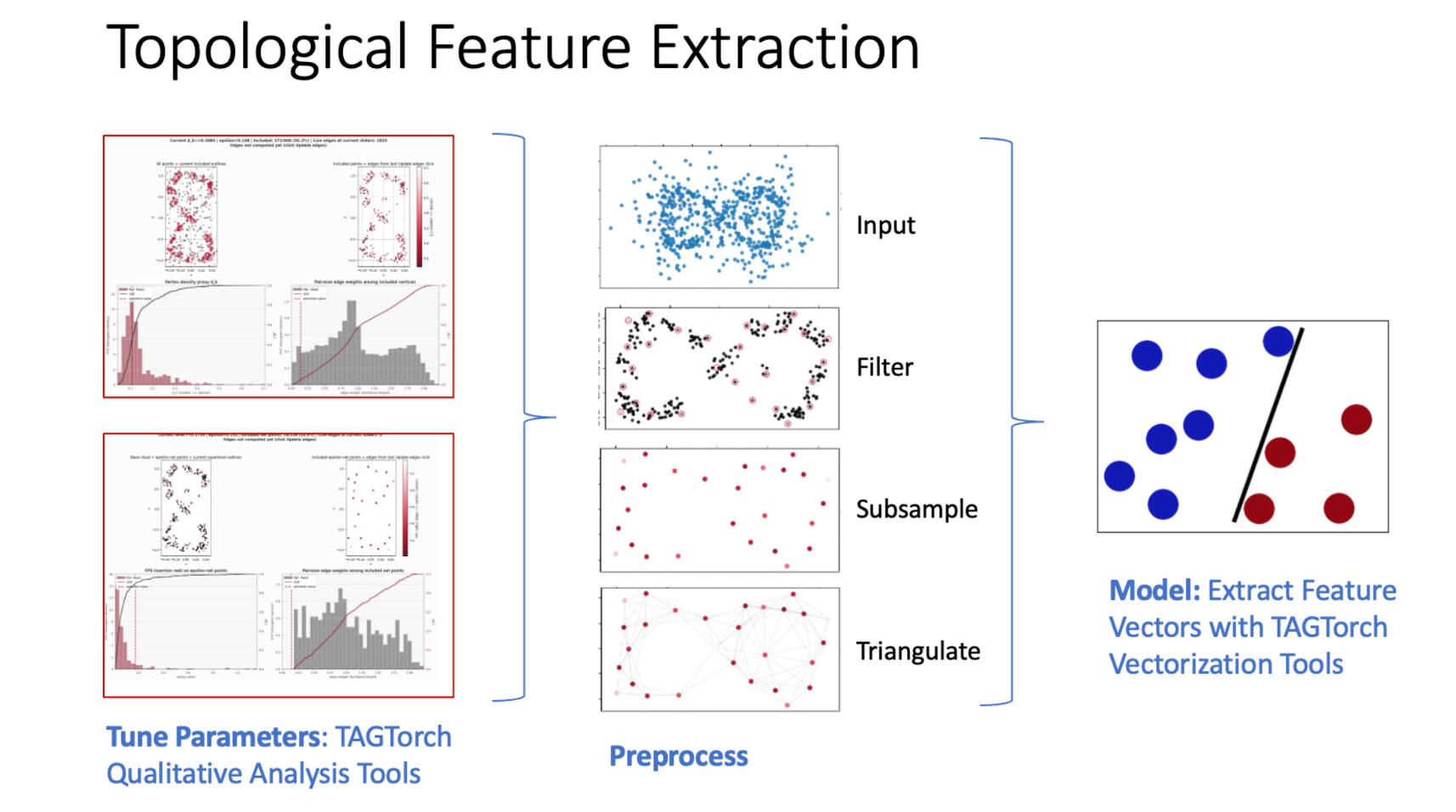}
    \caption{TAGTorch enables Human-in-the-Loop (HIL) workflows that are necessary to tune the process of topological data transforms.}
    \label{SM:fig:hyperparam-tuning-widget}
\end{figure}

\subsubsection{Building equivariant neural network models}

There is a diverse collection of strategies for building equivariance into neural networks, spanning data transformation and augmentation to designing structurally equivariant architectures. \tagtorch\; is designed to provide a unified access point to tools and model definitions in other packages in the community and implements other architectures. But defining networks is only half the battle --- significant domain knowledge and practical expertise, gained through years of familiarity with the research topic and underlying mathematics, make configuring models complex, often requiring experimentation and matching model design and specification to data types and modeling tasks.

As it matures, \tagtorch\; will support model specification through serialized configs (e.g., JSON or YAML files), defining model sizes, layer types, and preprocessing. Doing so will allow internal and community contributors to add configurations that define ``recipes'' for building networks specifically suited to certain data modalities and modeling objectives.

The other half of TAG and deep learning is training models. As with all model training, this requires expertise with hyperparameter tuning, tricks of the trade, and knowing which metrics and indicators of a well-trained model. Training workflows in \tagtorch\; will integrate with established training libraries and utilities in PyTorch, as well as specialized hyperparameter tuning software, pairing equivariant architectures and training procedures with traditional deep learning infrastructure.

\end{document}